# Deep Reinforcement Learning-Based Long-Range Autonomous Valet Parking for Smart Cities

Muhammad Khalid, Liang Wang, Kezhi Wang, Cunhua Pan, Nauman Aslam and Yue Cao

*Abstract*—In this paper, to reduce the congestion rate at the city center and increase the travelling quality of experience (QoE) of each user, the framework of long-range autonomous valet parking (LAVP) is presented, where an Autonomous Vehicle (AV) is deployed can pick up, drop off users at their required spots, and then drive to the car park around well organized places of city autonomously. In this framework, we aim to minimize the overall distance of the AV, while guarantee all users are served with great QoE, i.e., picking up, and dropping off users at their required spots through optimizing the path planning of the AV and number of serving time slots. To this end, we first propose a learning based algorithm, which is named as Double-Layer Ant Colony Optimization (DL-ACO) algorithm to solve the above problem in an iterative way. Then, to make the real-time decision, while considers the dynamic environment (i.e., the AV may pick up and drop off users from different locations), we further present a deep reinforcement learning (DRL) based algorithm, which is known as deep Q-learning network (DQN). The experimental results show that the DL-ACO and DQN-based algorithms both achieve the considerable performance.

## I. INTRODUCTION

The mobility of urban area is highly dependent on the transportation system. Effective transportation plays an important role in the sustainability and development of future smart cities [1]. Normally, city centers are the busiest places and difficult for the traffic congestion control [2]. A large number of vehicles enter and leave the city center every hour [3], [4]. It may result in various mobility issues, e.g., high congestion, pollution and fuel consumption rate as well as long journey time for the people. One of the key issues leading to the above problems may be that the vehicles keep searching for Car Park (CP) in city center [5]. According to the survey, 30% of overall traffic in urban area is caused by drivers searching for a suitable CP [6]. On average, it takes a driver about 6-20 minutes in UK to find a CP [7]. This is due to the fact that the drivers looking for CP may not have the prior and background information about CP or information about best route [8], [9]. To reduce the the searching and roaming time, the Smart Parking (SP) has been proposed, which gives drivers an opportunity to receive information of CP on their smart phones. The message provides user with real-time available parking slots in CP [10]. Thanks to the Information and Communication Technologies (ICT) as well as machine learning and computer vision based solutions, finding available CP slot becomes more and more convenient [11]. However, the SP may suffer from challenging scenarios where every vehicle rushes to the same CP and thus may result in a higher congestion rate in the city center [12], [13].

Almost all Autonomous Vehicle (AV) operates on battery power and can move autonomously, contributing a lot to eco-friendly environment. AV can reduce the carbon emission and minimize journey cost when compared with fuel-powered cars [14]. In addition, AV is expected to offer Autonomous Valet Parking (AVP), which can help the AV to find the suitable car park slot in CP. Normally, AVP provides two kinds of solutions: 1) Short-range Autonomous Valet Parking (SAVP); and 2) Long-range Autonomous Valet Parking (LAVP).

For the first case, the user can leave vehicle at CP entrance. Then SAVP can guide the AV to find a suitable parking spot itself inside CP. Specially, the AVP applies 3-dimensional (3D) localization and computer vision techniques to move between different stories of a CP. It also searches for a vacant parking slot as well as avoid obstacles on their way [15]. In addition, users can pick up their AV at the entrance of CP for the return journey, which may be synchronized with the movement of user to avoid delays as well as congestion inside CP. However, the key issue of SAVP is that the users still have to go a long way to the entrance of CP themselves, probably from their work place or city center via other means, however, as mentioned earlier, this process may take around 6-20 minutes on average. On the other hand, the LAVP proposes to deploy CP at the border of the city center to avoid congestion [16] inside the city center. In this case, instead of going to entrance of CP, the users can leave the AV anywhere. Such as the place close to their company or inside the city center where they can do shopping. We call these places as the drop-off spot. Once people gets out of the AV, then the AV can move to the available CP autonomously controlled by LAVP. For the return journey, users may use their mobile phone to book the AV, which will pick-up them at any spots as set by them, via applying the similar techniques. Normally, the drop-off spots can be set the same as the pick-up spots. To facilitate this process, the advanced vision and optimization techniques can

Muhammad Khalid is with the Department of Computer Science and Technology, University of Hull, Hull, HU6 7RX, U.K., email: m.khalid@hull.ac.uk.

Liang Wang, Kezhi Wang and Nauman Aslam are with the Department of Computer and Information Science, Northumbria University, Newcastle upon Tyne, NE1 8ST, U.K., emails: {liang.wang, kezhi.wang, nauman.aslam}@northumbria.ac.uk.

Cunhua Pan is with School of Electronic Engineering and Computer Science, Queen Mary University of London, E1 4NS, U.K., email: c.pan@qmul.ac.uk.

Yue Cao (corresponding author) is with School of Cyber Science and Engineering, Wuhan University, Wuhan 430072, China. Email: 871441562@qq.com

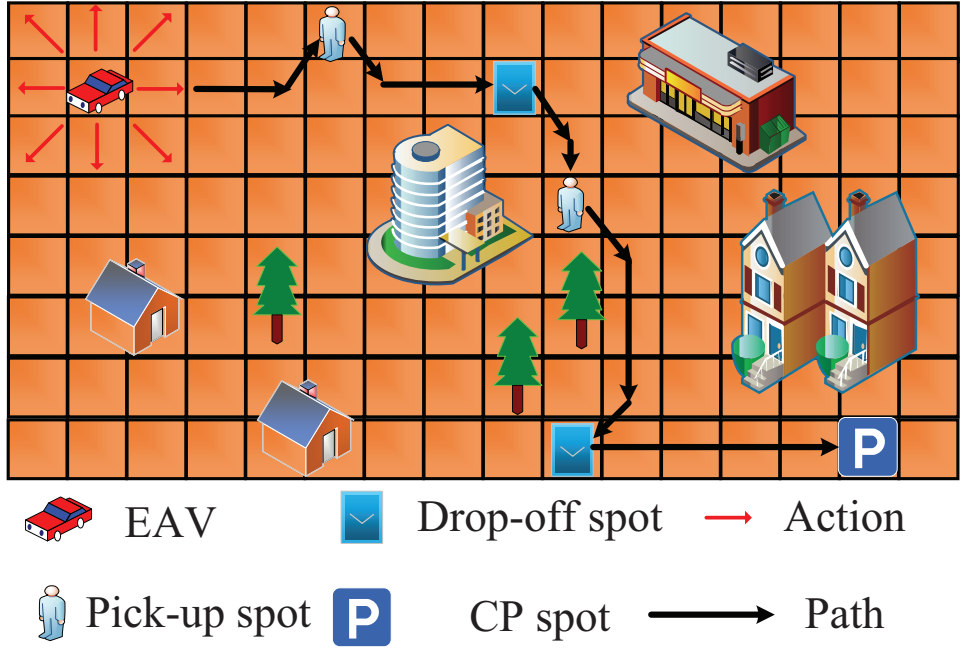

Fig. 1: The proposed long-range autonomous valet parking framework.

be applied to minimize overall cost (e.g., resources, battery, time) of AVs during the above process.

In above mentioned scenarios, the key challenges are yet to be addressed, i.e., path planning of the AV in dynamic environment, while also consider 1) meeting the quality of experience (QoE) of each user, 2) serving the user with the optimal order, i.e., picking up user first and dropping off it later, 3) minimizing the overall distance of the AV. The above optimization normally includes the mixed integer variables, which is very difficult to tackle. The traditional convex-based solutions, due to high complexity, cannot be applied in the above situations. Additionally, the random search does not normally provide optimal solution, as it is difficult to converge due to the non-learning and environment adaptability behavior. Furthermore, other hybrid optimization based solutions, e.g., greedy optimization normally needs several iterations and also suffer from high complexity. Also, if the environment varies, the above mentioned optimizations all have to be re-run to obtain the new solutions, which may not suitable for the real-time decision making in the dynamic environment.

Against the above background, in this paper, we aim to design the long-range autonomous valet parking framework, as shown in Fig. 1.

In this framework, we optimize the path planning of the AV, aiming at minimizing the overall distance of the AV, while guaranteeing all users are served, by picking up, dropping off them to their required spots. To this end, we propose two learning based solutions, i.e., Double-Layer Ant Colony Optimization (DL-ACO) and DQN based algorithm to solve the above problem. The DL-ACO can be applied to the new scenario or unfamiliar environment, especially the case when DQN is difficult to converge. The DL-ACO is also used to verify the results of DQN based algorithm in uncertain scenarios. On the other hand, the DQN can be applied to the familiar environment to make the efficient and real-time decision. This is because we can pre-train DQN and once the training is done, DQN can make decision very fast.

The remainder of this paper is organized as follows. Section II presents the related work and Section III describes the system model. Section IV introduces the proposed DL-ACO algorithm, while Section V presents the DQN-based algorithm. The simulation results are reported in Section VI, followed by conclusions in Section VII.

## II. LITERATURE REVIEW

Finding a suitable car parking slot is one of the key challenges in transportation systems. Optimal path planning plays an important for vehicles to search for the suitable car park slot while also minimizing the resources required, e.g., battery consumption [17]. Popular path planning and route optimization algorithm includes Dijkstra, Ant Colony Optimization (ACO) and A-Star (A*) [18], which can help to find the shortest path from source to destination. The Dijkstra algorithm divides road into edges and each edge is assigned with the weight. The weight of edges varies from scenario to scenario, which can either be energy, time or distance. In ACO, the artificial ants use swarm intelligence to find the shortest route. The biological ants when searches for food can leave some hormones called pheromone on their way to destination. This pheromone is sensed by other ants and then they can follow the same path. If more ants travel with the same path, there will a higher amount of pheromone. However, the above-mentioned algorithms do not normally learn from the past experience. This means if the source or destination changes, they may have to re-solve the problem to get the new optimal solutions.

Machine learning based solution has been proposed recently to assist real-time decision making in various situations. It mainly involves three main subcategories: supervised learning, unsupervised learning and reinforcement learning [19]. Supervised learning requires labelled data for decision making, whereas unsupervised learning is applied to learn data pattern and relationships from un-labelled or unknown data. The reinforcement learning, on the other hand, can be applied to the environments where no prior information is available. The learning process is achieved through direct interaction with environment. The reinforcement learning normally has five main elements: an agent, environment, state, action and reward [20]. In reinforcement learning, an agent can interact with the environment by taking actions and obtain the accumulated reward. The reinforcement learning may contain multiple episodes to fully train an agent in the environment. Then the decision making can be done, with the help of the agent. The reinforcement learning technique has shown huge potentials in autonomous vehicles, such as Unmanned Aerial Vehicles (UAVs). In [21], Wang *et al.* proposed a RLAA algorithm based on Q-learning to optimize the user association and resource allocation in UAV-enabled MEC. In [22], an UAV-aided MEC framework is investigated, where a group of UAVs cooperate to serve ground UEs, and the authors developed a MAT algorithm based on multi-agent reinforcement learning to optimize the trajectories of UAVs.

In this paper, to assist LAVP in the dynamic environment, two learning-based algorithms are proposed. Specifically, the DL-ACO algorithm can be applied in unfamiliar scenarios, whereas DQN-based algorithm can be deployed in dynamic or familiar environment, after adequate training.

## III. System Model

In this section, we consider the LAVP scenario, as shown in Fig. 1, where the city map is divided into a grid map. We assume that there is an AV serving $N$ users within the city. To simplify, we divide the city into a $Z^X \times Z^Y$ grid map, which contains several obstacles, and we define the set of users as $\mathcal{N} \triangleq \{n = 1, 2, ..., N\}$. The AV starts to serve users from the initial taking off spot, whose coordinate is denoted as $q^I = [X^I, Y^I]$. Additionally, the AV serves the user $n$ by visiting the pick-up spot, whose coordinate is $q_n^P = [X_n^P, Y_n^P]$, and visiting the drop-off spot, whose coordinate is $q_n^D = [X_n^D, Y_n^D]$. Finally, the AV reaches the CP spot, whose coordinate is denoted as $q^C = [X^C, Y^C]$, after all users are served.

Note that we assume this process lasts for $T$ time slots or steps, which vary with the path planning of AV. For simplicity, we denote the set of time slots as $\mathcal{T} \triangleq \{t = 1, 2, ..., T\}$, and in the time slot $t$, the AV will select an action $a(t)$ from the action set, which is denoted as $\mathcal{A} \triangleq \{$UP, DOWN, LEFT, RIGHT, TOP-LEFT, TOP-RIGHT, BOTTOM-LEFT, BOTTOM-RIGHT$\}$. Thus, it has

$$a(t) \in \mathcal{A}, \forall t \in \{1, 2, ...T\}. \quad (1)$$

Then, given the coordinate of the AV, denoted by $q(t) = [X(t), Y(t)]$ in time slot $t$, the coordinate of the AV $q(t+1) = [X(t+1), Y(t+1)]$ in next time slot is defined as follows:

$$q(t+1) = \begin{cases} [X(t)-1, Y(t)], & \text{if UP} \\ [X(t)+1, Y(t)], & \text{if DOWN} \\ [X(t), Y(t)-1], & \text{if LEFT} \\ [X(t), Y(t)+1], & \text{if RIGHT} \\ [X(t)-1, Y(t)-1], & \text{if TOP-LEFT} \\ [X(t)-1, Y(t)+1], & \text{if TOP-RIGHT} \\ [X(t)+1, Y(t)-1], & \text{if BOTTOM-LEFT} \\ [X(t)+1, Y(t)+1], & \text{if BOTTOM-RIGHT} \end{cases} \quad (2)$$

Thus, the distance traversed between $t$ and $t-1$ is expressed as

$$d(t, t-1) = \sqrt{||q(t) - q(t-1)||^2}, \ \forall t \in \mathcal{T}, \quad (3)$$

where $||\cdot||$ denotes Euclidean norm. Also, the AV can only move to its adjacent grid within the target grid map in each time slot. It has:

$$0 \leq X(t) \leq Z^X, \ \forall t \in \mathcal{T}, \quad (4)$$

and

$$0 \leq Y(t) \leq Z^Y, \ \forall t \in \mathcal{T}. \quad (5)$$

In each time slot, the AV will also need to avoid obstacles, and it has

$$q(t) \notin O, \forall t \in \mathcal{T}, \quad (6)$$

where $O$ is the set of the coordinates of obstacles.

In our paper, the travelling of AV is associated with pick-up spot, whose coordinate is $q_n^P = [X_n^P, Y_n^P]$, and a drop-off spot, whose coordinate can be expressed as $q_n^D = [X_n^D, Y_n^D]$. For simplicity, we assume if the AV reaches the pick-up spot where user $n$ is present, the user is picked by the AV, which can be expressed as $q(t) = q_n^P$, i.e., $[X(t), Y(t)] = [X_n^P, Y_n^P]$. Similarly, if the AV reaches the drop-off spot, it drops that user $n$ at that spot, which can be defined as $q(t) = q_n^D$, i.e., $[X(t), Y(t)] = [X_n^D, Y_n^D]$. Without loss of generality, we introduce the set $U(t)$ to present the serving status of all users in time slot $t$ as follows:

$$U(t) = \{u_n(t), \forall n \in \mathcal{N}\}, \forall t \in \{1, 2, ..., T\}, \quad (7)$$

where $u_n(t)$ is the serving status of user $n$ in time slot $t$, which has

$$u_n(t) = \begin{cases} 0, & \text{if } q_n^P \notin \{q(t'), t' = 1, 2, ..., t\}, \\ & \text{and } q_n^D \notin \{q(t'), t' = 1, 2, ..., t\}, \\ 1, & \text{if } q_n^P \in \{q(t'), t' = 1, 2, ..., t\}, \\ & \text{and } q_n^D \notin \{q(t'), t' = 1, 2, ..., t\}, \\ 2, & \text{if } q_n^P \in \{q(t_1), t_1 = 1, 2, ..., t\}, \\ & \text{and } q_n^D \in \{q(t_2), t_2 = 1, 2, ..., t\}, t_1 < t_2, \end{cases} \quad (8)$$

in which, from the first time slot to the current time slot $t$, 1) if the AV does not reach the pick-up and drop-off spot,

$u_n(t) = 0$; 2) if the AV reaches the pick-up spot, but does not reach the drop-off spot, $u_n(t) = 1$; and 3) if the AV first reaches the pick-up spot, and then arrives the drop-off spot, $u_n(t) = 2$.

After all the users are served, the AV then reaches the CP spot, and serving process terminates. The serving process terminates. Thus, we can have:

$$\sum_{n=1}^{N} u_n(t) = 2N, t = T, \quad (9)$$

and

$$q(t) = q^C, t = T. \quad (10)$$

Besides, the AV will start serving users from the initial taking off spot. Thus, it has:

$$q(t) = q^I, t = 0. \quad (11)$$

We aim to minimize the overall distance for the AV, while at the same time to make sure all users are served, through optimizing the path planning of the AV. To this end, we formulate our problem as follows:

$$\mathcal{P} : \min_{q,T} \sum_{t=1}^{T} d(t, t-1) \quad (12a)$$

subject to Constraints :

$$(1), (4), (5), (6), (9), (10), (11),$$

in which, $q = \{q(t), \forall t \in \mathcal{T}\}$, and $T = \{1, 2, ..., T\}$.

The above optimization $\mathcal{P}$ is quite challenging to tackle as one has to decide the optimal path planning, i.e., $q$ and the minimal number of time slots, i.e., $T$ that the AV moves. Precisely, the AV has to avoid obstacles in each time slot, and it must serve users in the adequate order. That is to say each of user should be picked up first and then dropped off next to the adequate spot. After that the AV has to reach the CP spot.

Next, we will propose two solutions to address the above problem. Firstly, we propose a learning based DL-ACO algorithm, which can achieve the optimal solutions in an iterative way. The DL-ACO algorithm can be applied in some unfamiliar scenarios. Then, to adapt to the dynamic environment, we further present a DQN-based algorithm, which can achieve the optimal solutions in real-time, once the training process is done.

## IV. PROPOSD DL-ACO ALGORITHM

The ACO can be applied in different optimization scenarios such as graph traversing, job scheduling and travelling salesman problems. The basic idea is to apply the biological ants working in form of a group to find a global optimal path. This path is between their nest and destination. The biological ants can leave a special kind of chemical called pheromone when they search for the destination. Then, the following ants will search according to pheromone left on the ground. Besides, the density of pheromone is related to the overall distance between the nest and destination. Thus, if the density of pheromone on a particular path is the highest among others, Then, most ants choose this path, which is the shortest path.

In our paper, the DL-ACO algorithm consists of two parts. Specifically, we first apply it to achieve the optimal path planning between each pair of spots, including taking-off, pick-up, drop-off, and CP spot. Then, we further apply it to find the optimal order for serving users. The details are given as follows.

---

**Algorithm 1** Path planning between different spots

1: Initialize the distance matrix $\boldsymbol{D^m}$ with size $2N + 2$;
2: Establish the spot list $E^s$ with size $2N + 2$;
3: **for** spot $i \in E^s$ **do**
4:   **for** spot $j \in E^s$ **do**
5:     Initialize pheromone matrix $\boldsymbol{p^m}$ with size $Z$;
6:     Initialize vector $D^e$;
7:     **for** iteration $l = 1, 2, ...e^{max}$ **do**
8:       **for** ant $k = 1, 2, ...k^{max}$ **do**
9:         **for** $t = 1, 2, ..., T^{max}$ **do**
10:           Select an action from $\mathcal{A}$ with probability $p_{v,w}^k$;
11:         **end for**
12:         **if** ant $k$ reaches spot $j$ from spot $i$ **then**
13:           Calculate overall distance $d_k^a$ according to (2);
14:           $D^e \leftarrow d_k^a$;
15:           Update pheromone matrix $\boldsymbol{p^m}$;
16:         **end if**
17:       **end for**
18:     **end for**
19:     $D_{i,j}^m = \min(D^e)$;
20:   **end for**
21: **end for**

---

Motivated by the work in [23], the first part of the proposed DL-ACO is described in the Algorithm 1. Specifically, we first define a distance matrix $\boldsymbol{D^m}$ with size $2N + 2$, which is used to store the minimal distance between each pair of spots, as shown at Line 1. Besides, we establish a spot list $E^s$ with size $2N + 2$ that the AV must reach. Then, from Line 3, we start to find the minimal distance value between spot $i$ and spot $j$. From Line 5 to 6, we initialize the pheromone matrix $\boldsymbol{p^m}$ with size $Z$, and we define a vector $D^e$ to store the successful distance value achieved by each ant. Then, from Line 10, the ant $k$ starts to explore and selects an action from $\mathcal{A}$. Note that the ant $k$ moves from the current grid $v$ to the next available grid $w$ with the probability $p_{v,w}^k$, expressed as follows:

$$p_{v,w}^k = \frac{\tau_{v,w}^\alpha \eta_{v,w}^\beta}{\sum_{z \in \text{allowed}_v} \tau_{v,z}^\alpha \eta_{v,z}^\beta}, \quad (13)$$

where $\tau_{v,w}$, $\eta_{v,w}$ are the amount of pheromone left and the attractiveness from grid $v$ to grid $w$, $\eta_{v,w}$. $\alpha$, $\beta$ are influence parameters which determine the importance of pheromone versus heuristic information.

Then, from Line 12, if the ant $k$ reaches the spot $j$, the overall distance $d_k^a$ between the pair of spots can be obtained by (2), and will be stored in $D^e$. Additionally, according to the grids that the ant $k$ visits, the pheromone matrix $\boldsymbol{p^m}$ will be updated given the following equation

$$\tau_{v,w} \leftarrow (1-\rho)\tau_{v,w} + \sum_k \triangle \tau_{v,w}^k, \quad (14)$$

where $\rho$ denotes the pheromone decay and $\triangle \tau_{v,w}^k$ can be

obtained by

$$\triangle \tau_{v,w}^k = \begin{cases} \frac{\mu}{d_k^a}, & \text{if ant } k \text{ visits path between grid } v, w, \\ 0, & \text{otherwise} \end{cases} \quad (15)$$

where $\mu$ is the constant value.

Furthermore, from Line 19, after adequate iteration, we select the minimal distance value from $D^e$ and store it in $D^m$. Then the optimal path planning between spot $i$ and spot $j$ is achieved.

---

**Algorithm 2** DL-ACO algorithm for LAVP

1: Obtain the distance matrix $D^m$ from Algorithm 1;
2: Initialize the pheromone matrix $p^s$ with size $2N+2$;
3: Initialize the minimal distance value $d^{min}$;
4: **for** iteration $l = 1,2,...,l^{max}$ **do**
5:   **for** ant $k = 1,2,...k^{max}$ **do**
6:     Select spots from spot list $E^s$;
7:     Obtain overall distance $d_k$ according to $D^m$;
8:     **if** (9), (10), (11) are met and $d_k < d^{min}$ **then**
9:       Update pheromone matrix $p^s$;
10:       $d^{min} \leftarrow d_k$;
11:     **end if**
12:   **end for**
13: **end for**
14: Obtain optimal path planning according to $d^{min}$.

---

Next, having obtained the distance matrix $D^m$ that can represent the minimal distance between each pair of spots, we further achieve the optimal order for serving users. Precisely, we show the overall algorithm in Algorithm 2. From Line 2 to 3, we initialize the pheromone matrix $p^s$ with size $2N+2$. Besides, we temporarily define a value $d^{min}$ to represent the minimal overall distance value achieved by the ants. Then, from Line 4, each of ant starts to explore and selects the spot that it will visit. Specifically, the ant $k$ moves from the current spot $i$ to the next available spot $j$ with the probability $p_{i,j}^k$, also presented in Fig. 2.

$$p_{i,j}^k = \frac{\tau_{i,j}^\alpha \eta_{i,j}^\beta}{\sum_{z \in \text{allowed}_i} \tau_{i,z}^\alpha \eta_{i,z}^\beta}. \quad (16)$$

After all spots are visited, we obtain the overall distance $d_k$. If the order of spots that the ant $k$ visits can meet requirement of (9), (10), (11), and in the same time $d_k < d^{min}$ is met, we update the pheromone matrix $p^s$ as follows:

$$\tau_{i,j} \leftarrow (1-\rho)\tau_{i,j} + \sum_k \triangle \tau_{i,j}^k, \quad (17)$$

where $\triangle \tau_{i,j}^k$ is

$$\triangle \tau_{i,j}^k = \begin{cases} \frac{\mu}{d_k^a}, & \text{if ant } k \text{ visits path between spot } i, j, \\ 0, & \text{otherwise.} \end{cases} \quad (18)$$

Then, we set $d_k$ as the temporarily minimal distance value $d^{min}$. After adequate iteration, we can obtain the optimal path planning according to $d^{min}$.

## V. DQN-BASED ALGORITHM

The above DL-ACO algorithm may not be suitable for dynamic environment (where pick-up, drop-off and route will change in every episode) as it needs to iterate to find the optimal solutions, and re-run if the locations of pick-up, drop-off spots changes at each cycle. Motivated by this, we introduce a DQN-based algorithm that can achieve the optimal solutions in real-time. Next, we first introduce some background knowledge of deep reinforcement learning, including deep neural network (DNN), Q-value, and other fundamental elements.

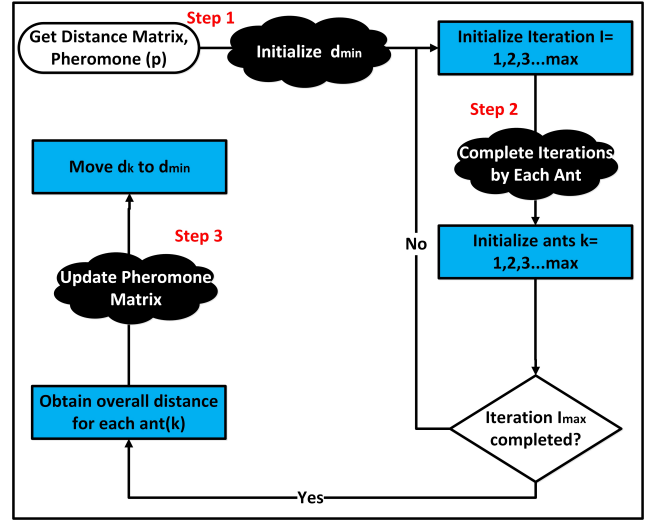

Fig. 2: Overall structure of DL-ACO for LAVP

### A. Background Knowledge

The reinforcement learning being an emerging technology playing a beneficial role in scenarios where environment is dynamic and keep changing frequently [24]. A general reinforcement learning model consist of an agent, action list, reward, states and environment [25]. Specifically, in the structure of reinforcement learning, an agent is considered to interact with the environment. The process of the interaction can be expressed as the finite Markov decision process (MDP). Specifically, given a series of states $s(t)$, the goal of the agent is to select actions $a(t)$ that can maximize the accumulated rewards $\sum_{t'=t}^{T} \gamma^{t'-t} r(t')$, where $\gamma$ is the discount factor that balances the immediate and future reward, $r(t)$ is the reward. Additionally, in order to map the relationship between state $s(t)$ and action $a(t)$, an action-value function, it is also known as Q-value $Q(s(t), a(t))$ is defined, which can be expressed as an Bellman equation. Besides, in order to obtain the accumulated rewards, another important element of reinforcement learning named Q-table [26] is applied to store the Q-value of each pair of action and state.

However, the classical reinforcement learning may suffer from high-dimensional space of states and actions as the size of Q-table is finite. Motivated by the development of deep neural networks (DNNs), Mnih *et al.* [27], [28] combined reinforcement learning and DNNs, i.e., DQN, which can replace the Q-table. Additionally, in order to further stabilize the training process, two mechanisms are proposed. First, the memory named experience replay is used to store the experiences of the past, which eases the correlation between each of states. When the action is generated by the DQN, the agent sends the action to the environment, and the state will transfer to the next state. Then, the experience, which consists of $[s(t), a(t), r(t), s(t + 1)]$ will be stored in the experience replay memory. When the learning process starts, several experiences will be sampled for training the DQN. The second mechanism is called target network, which has the same structure as the DQN, but it only updates with certain intervals.

### B. DQN-based algorithm for LAVP

In order to apply DQN-based algorithm in the LAVP framework, we define the action, state, and reward function as follows:

- **Action:** In our proposed scheme, we define $a(t)$ in each time slot as the action of the AV, and it has

$$a(t) \in \mathcal{A}. \quad (19)$$

- **State:** In our paper, the state $s(t)$ consists of the following components:
  - the current coordinate of the AV: $[X(t), Y(t)]$.
  - the coordinates of all pick-up spots: $[X_n^P, Y_n^P], \forall n \in \mathcal{N}$.
  - the coordinates of all drop-off spots: $[X_n^D, Y_n^D], \forall n \in \mathcal{N}$.
  - the coordinate of CP spot: $[X^C, Y^C]$.
  - the serving status of each user: $u_n(t), \forall n \in \mathcal{N}$.

- **Reward:**
  In order to achieve a better performance in terms of convergence, we define the reward function $r(t)$ as follows:

$$r(t) = \begin{cases} -p, & \text{if (4) - (6) are not met} \\ 2p, & \text{if } q(t) = q_n^P \text{ and } u_n(t) = 1 \text{ in the first time} \\ 4p, & \text{if } q(t) = q_n^D \text{ and } u_n(t) = 2 \text{ in the first time} \\ 10p, & \text{if } q(t) = q^C \text{ and } \sum_{n=1}^{N} u_n(t) = 2N \\ -d(t-1, t), & \text{otherwise,} \end{cases} \quad (20)$$

From above, it is observed that 1) if the AV crosses the border or hits the obstacle, the agent will obtain a reward, which is $-p$; 2) if the AV reaches the pick-up spot, i.e., $u_n(t) = 1$, and $q(t) = q_n^P$ in the first time, the reward is $2p$; 3) if the AV reaches the drop-off spot of user $n$ and reached the pick-up spot of user $n$ before, that is to say $u_n(t) = 2$, and $q(t) = q_n^D$ the agent will also receive a reward of $4p$; 4) if the AV reaches the CP spot and all users are served, which means (9), (10), (11) are met, the reward is $10p$; and 5) otherwise the reward is defined as the minus of distance $d(t-1, t)$.

We provide the overall structure of DQN-based algorithm for LAVP in Fig. 3. More precisely, the agent is deployed to control the AV through interacting with the environment, the agent sends the state $s(t)$ to the DQN named evaluation network, which generates the Q-values of all actions. Note that in order to avoid the local optimum, an $\epsilon$-greedy policy is applied. Then, the agent sends the action $a(t)$ to the environment and obtains the reward $r(t)$. After that, the environment transfers to the state $s(t+1)$ of the next time slot. The experience, which consists of $[s(t), a(t), r(t), s(t+1)]$, is stored in the experience replay memory $M$. When the learning process starts, $K$ experiences will be randomly sampled for obtaining the target value, which can be expressed as

$$Q^*(s(t), a(t)) = \mathbb{E}\left[r(t) + \gamma \max_{a(t+1)} Q^*(s(t+1), a(t+1); \theta^-)\right], \quad (21)$$

where $\gamma \in [0, 1]$ is the discount factor $Q^*$ is generated by the target network, $\theta^-$ denotes the network parameter.

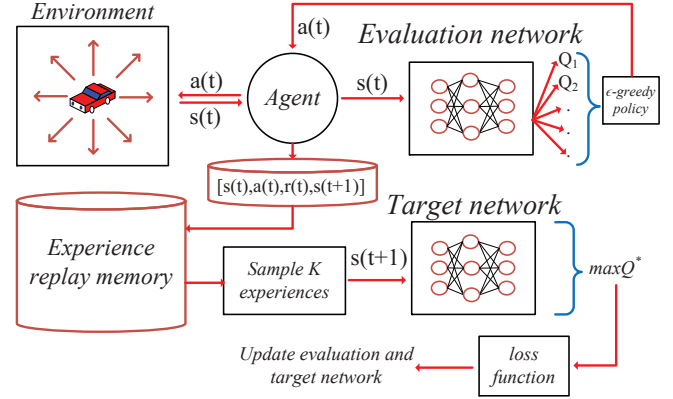

Fig. 3: Overall structure of DQN for LAVP

Then, the loss function can be calculated as follows:

$$L(\theta) = \mathbb{E}\left[\left(Q^*(s(t), a(t)) - Q(s(t), a(t); \theta)\right)^2\right]. \quad (22)$$

The evaluation network is updated according to the following equation:

$$\nabla_\theta L(\theta) = \mathbb{E}\left[\left(Q^*(s(t), a(t)) - Q(s(t), a(t); \theta)\right) \nabla Q(s(t), a(t); \theta)\right]. \quad (23)$$

Furthermore, we show the overall algorithm design in Algorithm 3, from which, we first initialize the evaluation network, and target network with parameters $\theta$, $\theta^-$ respectively at Line 1. The experience replay memory $M$ is also initialized at Line 2. Then, in each training episode, we initialize the state in the first time slot. The agent interacts with the environment given the state $s(t)$ and receive the action from the evaluation network. Note that for avoiding the local optimum, an $\epsilon$-greedy policy is applied, which means the agent can obtain the action that has the largest Q-value with probability $\epsilon$, or randomly obtains the action from $\mathcal{A}$ with probability $1 - \epsilon$. From Line 9, it obtains the reward according to (20), and transfers to the state of next time slot $s(t + 1)$. Then, the experience $[s(t), a(t), r(t), s(t + 1)]$ is stored into the experience replay memory. From Line 12, if the learning process starts, $K$ experiences will be randomly sampled for training the networks. Specifically, we obtain the loss value from (22), and train the evaluation network from (23). Additionally, the target network will be updated with the rate $\tau$.

**Algorithm 3** DQN-based algorithm for LAVP

1: Initialize evaluation and target network with parameter $\theta$, $\theta^-$;
2: Initialize experience replay memory $M$;
3: **for** $E = 1, 2, ..., E^{max}$ **do**
4:     Initialize state $s(t)$;
5:     **for** $t = 1, 2, ...T^{max}$ **do**
6:         Obtain state $s(t)$ from the environment;
7:         Select action $a(t)$ that has the largest Q-value with probability $\epsilon$;
8:         Randomly select action from $\mathcal{A}$ with probability $1 - \epsilon$;
9:         Obtain reward according to (20);
10:        Obtain the state of next time slot $s(t+1)$;
11:        Store experience $[s(t), a(t), r(t), s(t+1)]$ in memory;
12:        **if** learning process starts **then**
13:            Randomly sample $K$ experiences from memory;
14:            Calculate loss function according to (22);
15:            Train the evaluation network according to (23) with learning rate $\phi$;
16:            Update parameters of the target network: $\theta^- = \tau \theta^- + (1 - \tau)$;
17:        **end if**
18:     **end for**
19: **end for**

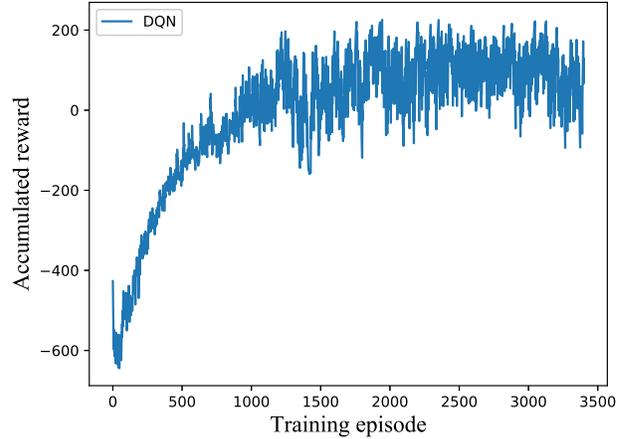

Fig. 4: The convergence performance of DQN.

## VI. Performance Analysis

The simulation has been carried out in Python 3.7, tensorflow 1.15.0, INTEL 3450T, and NVIDIA GTX 1050Ti. We divide the simulation area as a $20 \times 20$ grid map, which consists of various roads, obstacles and buildings. The AV always starts to serve users from the initial taking off spot, whose coordinate is $q^i = [0, 0]$. After serving all users, the AV will move to the CP spot, whose coordinate is $q^c = [19, 19]$. Once the DQN training is completed, the AV can move to CP anytime. For simulation and analysis purposes, we instruct the AV to park at CP once all passengers are served. In our DQN-based algorithm, we deploy three fully-connected layers with [400, 300, 300] neurons. The AdamOptimizer [29] is used. The network is trained with the learning rate $\phi = 0.003$, the target network is updated with the rate $\tau = 0.001$, and the size of experience replay memory is $10^6$. More parameters can be found in Table. I.

TABLE I: Simulation Parameters

| Parameter | Description | Parameter | Description |
|---|---|---|---|
| $Z^X$ | 20 | $Z^Y$ | 20 |
| $q^i$ | [0, 0] | $q^c$ | [19, 19] |
| $N$ | 3 | $\alpha$ | 1.1 |
| $\beta$ | 12 | $\rho$ | 0.5 |
| $\mu$ | 10 | $e^{max}$ | 10 |
| $k^{max}$ | 20 | $l^{max}$ | 50 |
| $\phi$ | 0.0003 | $\gamma$ | 0.99 |
| $p$ | 10 | $\epsilon$ | 0.9 |
| $\tau$ | 0.001 | $E^{max}$ | $10^6$ |
| $T^{max}$ | 100 | $K$ | 256 |

We first analyze the convergence performance of DQN-based algorithm during the training process in Fig. 4, where there are 3 users waiting to be served by the AV. As shown in Fig. 4, we observe that the accumulated reward achieved by DQN remains at −600 at the beginning. The plausible explanation is that the neural network is not convergent and the AV always moves out of the grid map, which means the penalty is always incurred. After that, the accumulated reward increases rapidly, which means the network starts to converge. Then, after 1000 training episodes, the curve remains between 0 and 200 eventually, which means the DQN-based algorithm has achieved the optimal path planning.

Then, we depict the path planning of AV achieved by the proposed DL-ACO algorithm in Fig. 5. Note that the black rectangle represents obstacle, the gray rectangle represents the spot. In addition, $IS$ represents the initial taking off spot, $PS$ denotes pick-up spot, $DS$ is drop-off spot, $CP$ means CP spot, and red line is the path planning of AV. In Fig. 5(a), there are 3 users, the coordinates of their pick-up and drop-off spots are $q_n^p = \{[3, 4], [7, 9], [10, 5]\}$, $q_n^d = \{[14, 7], [17, 16], [15, 12]\}$ respectively. One can observe that the AV controlled by DL-ACO algorithm serves all users with the order $IS \rightarrow PS1 \rightarrow PS2 \rightarrow PS3 \rightarrow DS1 \rightarrow DS3 \rightarrow DS2 \rightarrow CP$. Additionally, we can also observe that in each pair of spots, the DL-ACO always achieved the shortest distance.

Then, in Fig. 5(b), we deploy another 3 users, whose coordinates of pick-up and drop-off spots are $q_n^p = \{[0, 6], [5, 6], [10, 8]\}$, $q_n^d = \{[15, 6], [15, 13], [19, 17]\}$ respectively. We can observe that the AV controlled by DL-ACO algorithm serves the users with the order $IS \rightarrow PS1 \rightarrow PS2 \rightarrow PS3 \rightarrow DS1 \rightarrow DS2 \rightarrow DS3 \rightarrow CP$. Besides, we can conclude that DL-ACO still achieved the shortest distance in each pair of spots.

Furthermore, we show the path planning of AV achieved by DL-ACO in another scenario, where the coordinates of users' pick-up and drop-off spots are $q_n^p = \{[4, 3], [13, 7], [18, 6]\}$, and $q_n^d = \{[14, 11], [12, 16], [16, 16]\}$. It is observed that the AV serves the users with the order $IS \rightarrow PS1 \rightarrow PS2 \rightarrow DS1 \rightarrow PS3 \rightarrow DS3 \rightarrow DS2 \rightarrow CP$.

Then, after adequate training, we save the network parameters of DQN for testing. We depict the path planning of AV controlled by DQN in Fig. 6. Note that the spots in Fig. 6 are the same as Fig. 5. As shown in Fig. 5(a), the AV serves the users with the order $IS \rightarrow PS1 \rightarrow PS2 \rightarrow PS3 \rightarrow DS1 \rightarrow DS3 \rightarrow DS2 \rightarrow CP$. The overall distance is the same as Fig. 5(a), although their path planning is slightly different.

In Fig. 5(b), the AV serves the users with the order $IS \rightarrow PS1 \rightarrow PS2 \rightarrow PS3 \rightarrow DS1 \rightarrow DS2 \rightarrow DS3 \rightarrow CP$, which

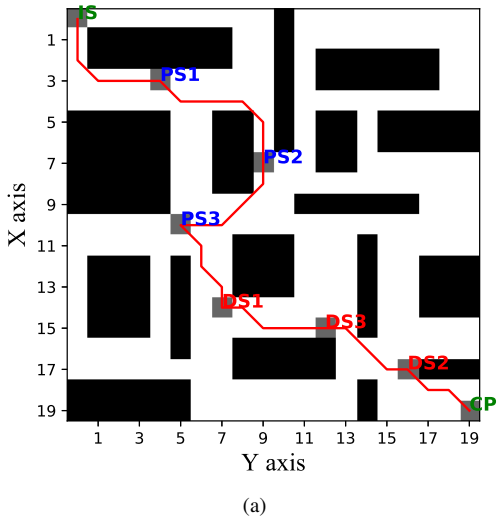

(a)

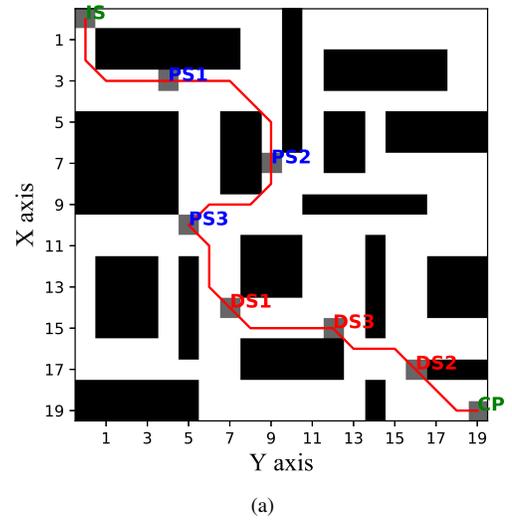

(a)

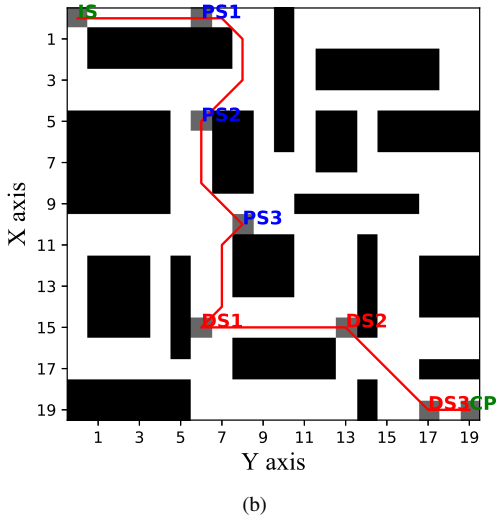

(b)

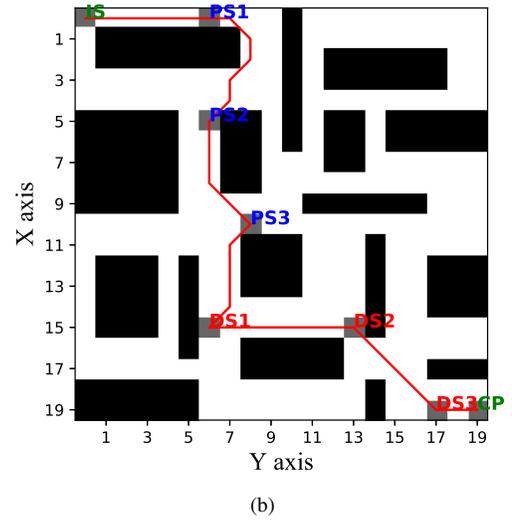

(b)

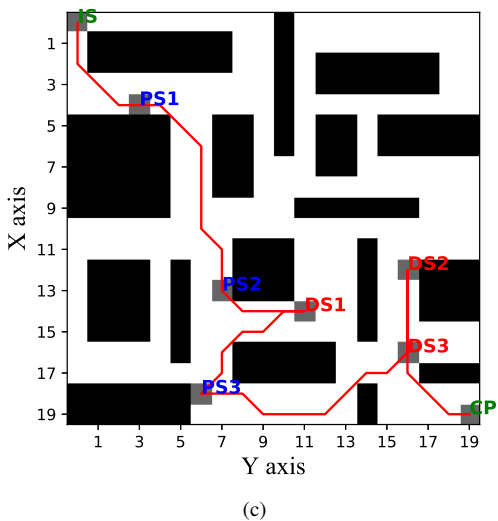

(c)

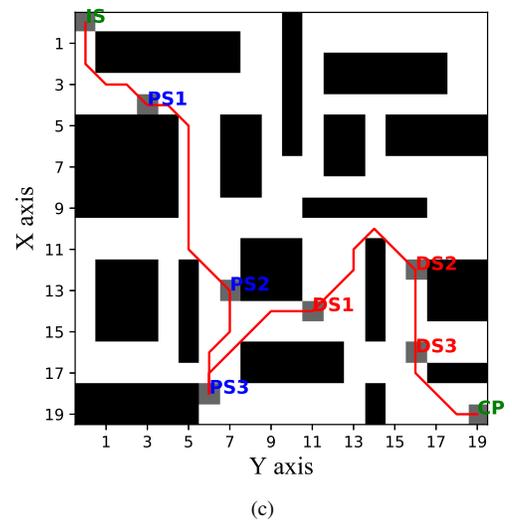

(c)

Fig. 5: The path planning of AV achieved by DL-ACO (IS for Initial Spot, PS for Pick-up Spot, DS for Drop-off Spot, CP for CP Spot)

Fig. 6: The path planning of AV achieved by DQN (IS for Initial Spot, PS for Pick-up Spot, DS for Drop-off Spot, CP for Car Parking Spot)

is the same as Fig. 5(b).

Then, as shown in Fig. 6(c), the AV serves the users with the order $IS \rightarrow PS1 \rightarrow PS2 \rightarrow PS3 \rightarrow DS1 \rightarrow DS2 \rightarrow DS3 \rightarrow CP$. Additionally, compared with the path planning in Fig. 5(c), DQN is much better as its' order for serving users is more practical, which will reduce unnecessary path.

Furthermore, we compare the execution time and distance achieved by DL-ACO, DQN in different cases, i.e., different locations of pick-up, drop-off spots, where there are 3 users. For DQN, the training episodes is 3500. In Table. II, we observe that no matter which case is, DQN always outperforms DL-ACO in terms of over distance. However, it is easy to see that DL-ACO consumes at least 100 seconds for achieving the considerable performance, while DQN only needs about 1 second in testing, although it takes longer time in training process.

TABLE II: Executed Time of DL-ACO and DQN

| Case | DL-ACO | | DQN | | |
|---|---|---|---|---|---|
| | Execution | Distance | Training | Testing | Distance |
| 1 | 112.99(s) | 38.971 | | 1.44(s) | 38.971 |
| 2 | 134.55(s) | 39.556 | | 2.40(s) | 39.556 |
| 3 | 203.66(s) | 53.213 | 1687.95(s) | 1.04(s) | 46.627 |
| 4 | 128.77(s) | 43.799 | | 0.93(s) | 40.385 |
| 5 | 138.35(s) | 38.971 | | 1.52(s) | 38.385 |

Finally, we analyze the performance of DL-ACO and DQN-based algorithm given different number of users. For fairness, we also apply another random algorithm as comparison. In this setting, the AV randomly selects a available action to take until all users are served and the AV reaches the CP spot. We obtain 500 pairs of path planning of AV achieved by Random and select the optimal result to compare with the proposed DL-ACO and DQN-based algorithms in Fig. 7, from which, we can observe that when the number of users are 1, 2, 4, and 5, our propsoed DL-ACO and DQN achieve the same performance in terms of overall distance of the AV. However, when the number of users is 3, DQN outperforms DL-ACO. One plausible explanation is that DL-ACO dose not find the optimal order for serving users, and this to longer path of the AV. While Random always performs the worst, no matter how many the number of users is.

## CONCLUSIONS

In this paper, we have presented a LAVP framework and proposed two learning based solutions, i.e., DL-ACO and DQN-based algorithms for minimizing the overall distance of the AV. In the meantime, guaranteeing all users are served, through optimizing the path planning and the number of time slots of the AV. The DL-ACO can be applied to new scenarios and maps pre-training is not required. The decision may take longer time and consume higher computation power as compared to the DQN-based algorithm. The DQN-based algorithm needs one time training. After the training is completed, this algorithm can take faster and efficient decision with less computation power than DL-ACO algorithm. The experimental results have shown that both DL-ACO and DQN-based algorithms can achieve the considerable performance, compared to other traditional algorithms. The DL-ACO and DQN-based algorithms can be easily extended to other urban transportation problems like electric vehicle charging, charging slot assignment and for unmanned-aerial vehicles used for various services.

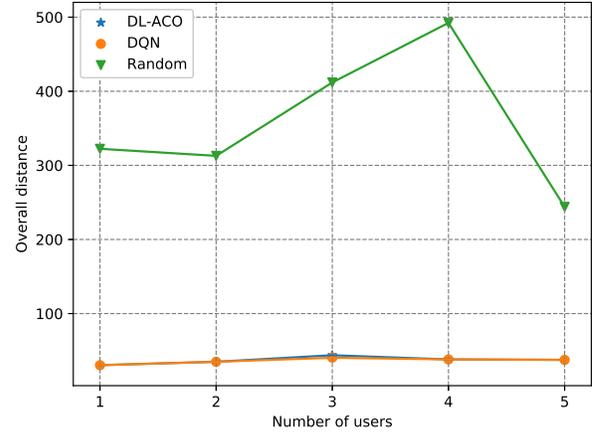

Fig. 7: The overall distance of AV achieved by DL-ACO, DQN and Random given different number of users.